\documentclass{egpubl}
\usepackage{eg2022}
 
\ConferencePaper        %

\usepackage[T1]{fontenc}
\usepackage{dfadobe}  
\usepackage{cite}  %

\BibtexOrBiblatex
\electronicVersion
\PrintedOrElectronic
\ifpdf \usepackage[pdftex]{graphicx} \pdfcompresslevel=9
\else \usepackage[dvips]{graphicx} \fi

\usepackage{egweblnk}

\usepackage{url}            %
\usepackage{booktabs}       %
\usepackage{amsfonts}       %
\usepackage{nicefrac}       %
\usepackage{microtype}      %
\usepackage{xcolor}         %
\usepackage{hyperref}       %

\usepackage{graphicx}
\usepackage{amssymb, bm}
\usepackage{amsmath}
\usepackage{siunitx}
\usepackage{mathtools,xparse}
\usepackage{xspace}
\usepackage{wrapfig}
\usepackage{multirow}
\usepackage{tabularx}
\usepackage{wrapfig}
\usepackage{sidecap}
\usepackage{subcaption}

\newcommand{\rev}[1]{{\color{black}#1}}

\title{
MoCo-Flow: Neural Motion Consensus Flow for Dynamic Humans in Stationary Monocular Cameras \\
}

\newcommand{\name}{MoCo-Flow\xspace}

\author[Chen, Li, Cohen-Or, Mitra, Chen]
{\parbox{\textwidth}{ \centering
        Xuelin Chen$^{1}$\hspace{.5cm}
        Weiyu Li$^{2,1}$\hspace{.5cm}
        Daniel Cohen-Or$^{3}$\hspace{.5cm}
        Niloy J. Mitra$^{4,5}$\hspace{.5cm}
        Baoquan Chen$^{6}$\hspace{.5cm}
        }
        \\
{\parbox{\textwidth}{ \centering
         $^1$Tencent AI Lab\hspace{.5cm}
         $^2$Shandong University\hspace{.5cm}
         $^3$Tel Aviv University\hspace{.5cm}
         $^4$University College London\hspace{.5cm}
         $^5$Adobe Research\hspace{.5cm}
         $^6$CFCS, Peking University
       } 
}
}

\begin{document}

\teaser{
    \centering
    \includegraphics[width=0.9\textwidth]{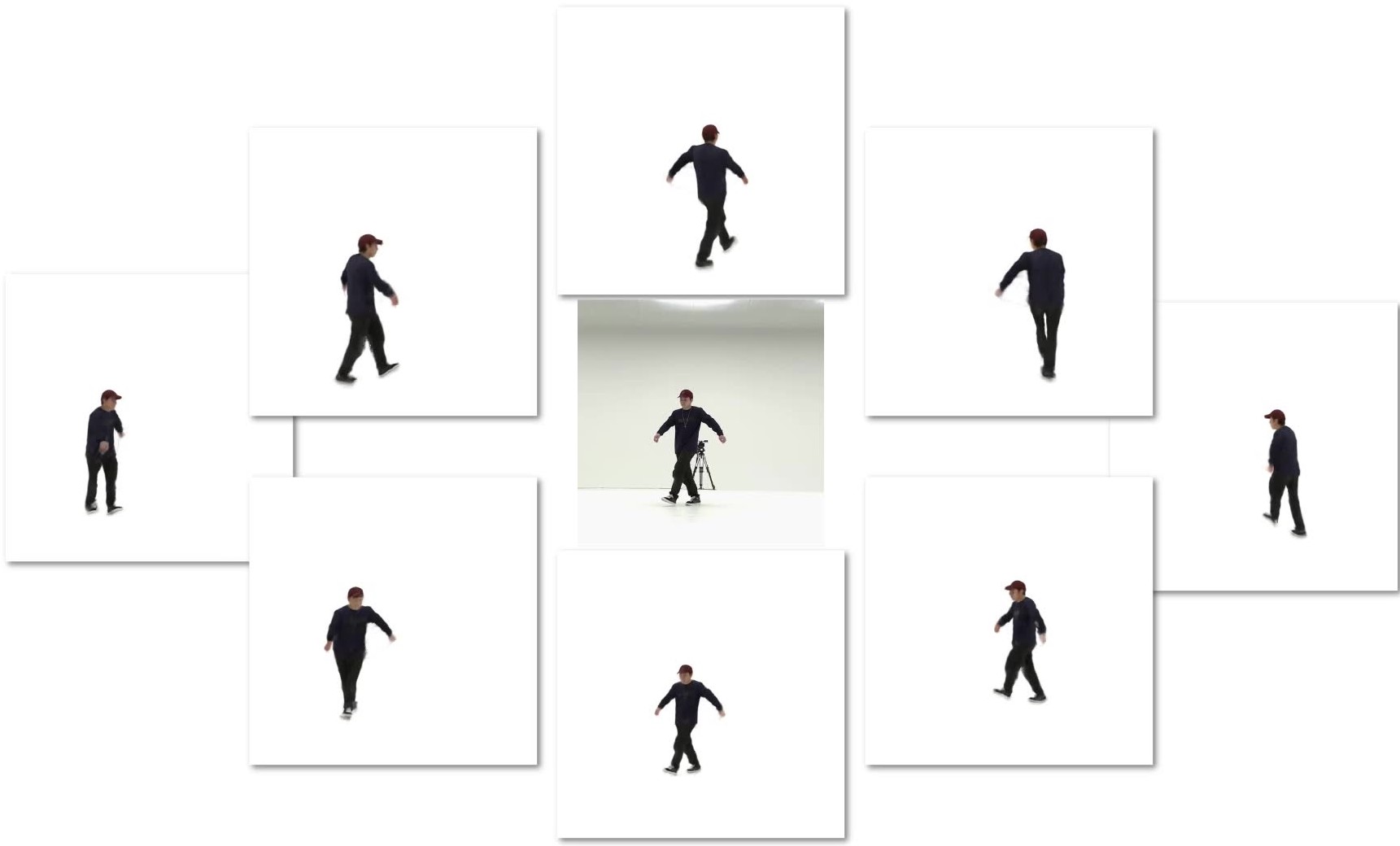}
    \caption{  
         Given only a video~(a) captured from a \textit{stationary monocular} camera, our method can synthesize novel views~(b) of the dynamic human from arbitrary viewpoints and at any time. An input view is shown in the middle on the right, with imagery synthesized from 8 novel views spreading around the performer. For dynamic results, please see the supplementary video. 
      }
\label{fig:teaser}
}

\maketitle

\begin{abstract}
Synthesizing novel views of dynamic humans from stationary monocular cameras is a specialized but desirable setup. This is particularly attractive as it does not require static scenes, controlled environments, or specialized capture hardware.
In contrast to techniques that exploit multi-view observations, 
the problem of modeling a dynamic scene from a single view is significantly more under-constrained and ill-posed. 
In this paper, we introduce Neural Motion Consensus Flow (\emph{\name}), a representation that models dynamic {humans in stationary monocular cameras} using a 4D continuous time-variant function.
We learn the proposed representation by optimizing for a dynamic scene that minimizes the total rendering error, over all the observed images.
At the heart of our work lies a carefully designed optimization scheme, which includes a dedicated initialization step and is constrained by a \emph{motion consensus} regularization on the estimated motion flow.
We extensively evaluate \name on several datasets that contain human motions of varying complexity, and compare, both qualitatively and quantitatively, to several baselines and ablated variations of our methods, showing the efficacy and merits of the proposed approach.
Pretrained model, code, and data will be released for research purposes upon paper acceptance. 

\ccsdesc[300]{Computing methodologies~Shape modeling}
\ccsdesc[300]{Computing methodologies~Rendering}

\printccsdesc   
\end{abstract}

\section{Introduction}

\begin{figure*}[t]
  \centering
  \includegraphics[width=0.9\textwidth]{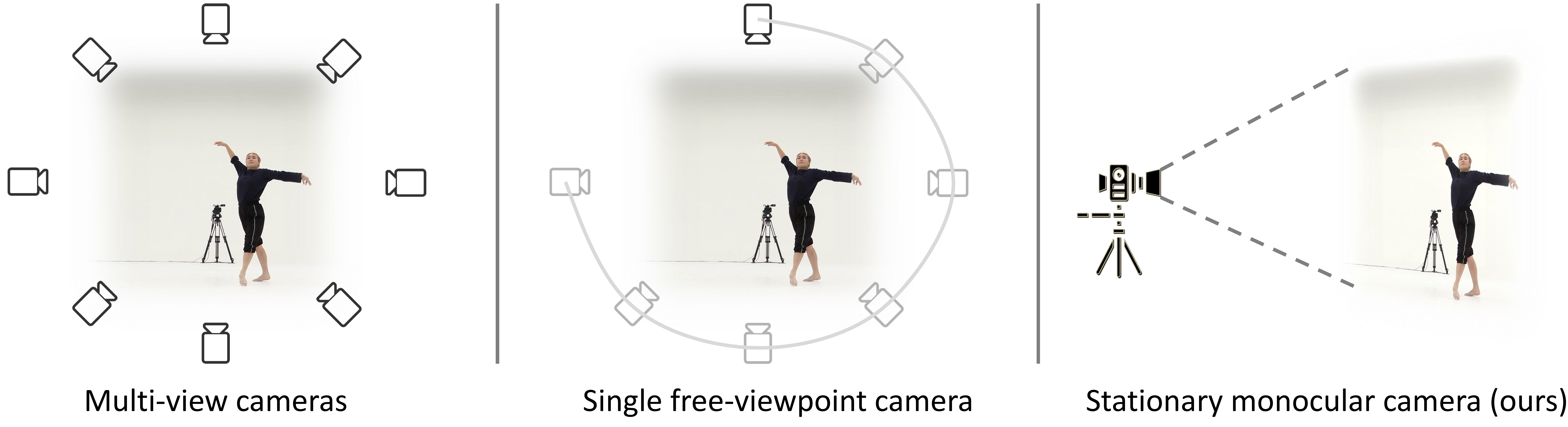}
  \caption{
  (Left)~Multi-view cameras setup for full observation of a dynamic scene; 
  (middle)~single free-viewpoint camera setup that captures the dynamics from varying viewpoints;
  (right)~stationary monocular camera  to observe a dynamic scene 
  from only a single \textit{fixed} viewpoint.
  }
  \label{fig:diff_settings}
\end{figure*}

We address the challenging problem of synthesizing novel views of dynamic humans in stationary monocular cameras.
View synthesis has been a long-standing problem in both computer vision and computer graphics.
Neural Radiance Field (NeRF)~\cite{nerf} has recently revolutionized novel view synthesis of \textit{static} structures by directly optimizing parameters of a continuous 5D scene representation to minimize the error of rendering multiple captured images.
Subsequently, there has been a surge of followups extending it to deal with dynamic scenes~\cite{nerfies, nsff, D-NeRF, video-nerf, neural3d, NerFACE, neuralbody, nonrigid_nerf, nerflow, park2021hypernerf}.

These dynamic NeRFs have shown impressive performance in view synthesis but require different setups to capture the dynamics.
The most natural extension is to still use a multi-view camera setting~\cite{neuralbody, neural3d} 
to acquire sufficient observation of the dynamic scene from multiple viewpoints.
While the multi-view setup significantly  constrains modeling of the dynamics, the capture process relies on controlled environments and specialized hardware to  synchronize the different acquisitions.  
Another line of work exploits a \emph{single} free-viewpoint camera to capture dynamic scenes from \emph{varying viewpoints}~\cite{nerfies, nsff, D-NeRF, video-nerf, nonrigid_nerf, nerflow}. 
While these methods bypass the need for expensive equipment, they still require the capture device to be suitably moved around to allow capturing dynamic scenes, and inevitably rely on Structure-from-Motion (SfM) systems for accurate camera extrinsic parameters to constrain the modeling.

In this paper, we present a dynamic NeRF technique for synthesizing novel views of dynamic humans from stationary monocular cameras.
The problem of modeling dynamic scenes from monocular cameras is typically under-constrained, as shown in~\cite{nerfies, nsff, D-NeRF, video-nerf, nonrigid_nerf, nerflow}. 
Moreover, our setting is even more challenging.
Unlike the multiple viewpoints setting, in our stationary monocular camera setting, we only observe the dynamic scene from a single fixed viewpoint. 
Hence, the extrinsic camera parameters cannot be obtained from SfM to constrain the dynamic scene modeling from multiple viewpoints as in aforementioned works (Figure~\ref{fig:diff_settings} illustrates different capture setups).
On the other hand, a technique for stationary monocular cameras has a significant scope and \rev{is} applicable to a wide range of everyday captures. 
\rev{Synthesizing} novel views from stationary videos would offer creating strong immersive experience for existing videos, and could be embraced in the future by millions of video content producers.
Furthermore, stationary videos are easy to capture and require no special assistance, environment or hardware.

In general, to model a dynamic scene for view synthesis, it can be decomposed into a shared canonical static scene for representing the appearance and the geometry of the subjects, 
and a motion flow that model the dynamics between the canonical space and the observation space at each frame.
Both representations can be approximated by Multi-layer Perceptron (MLP) networks and optimized to re-produce the observed frames via differentiable volume rendering.
However, a naive approach cannot deal with this overly ambiguous single fix-viewpoint setting, and the network update is prone to an erroneous overfit, 
as multiple solutions comprised of meaningless canonical representation and motion flows can be combined together to reproduce the observation. 

Hence, the key to solving the above formulation is to harness this challenging optimization, throughout the whole optimization process. 
This results in human bodies and their dynamics that follow faithfully the human perception of the video.
To this end, we devise a carefully designed and yet \emph{easy-to-implement} optimization scheme, 
which disambiguates relatively much worse local minima early at the initialization phase,
and imposes a crucial regularization on the update of the motion flow to reach a high degree of \emph{consensus} across the observations, denoted as Motion Consensus Flow (\name in short), consequently preventing the optimization from deviating too much from the initialization and landing on bad local minima.
It is worth noting that, in contrast to heuristic regularization terms, the regularization imposed in \name is general and does \emph{not} assume any dynamic characteristics on the moving subjects.

We demonstrate our method on several publicly available datasets, namely AIST~\cite{aist}, People-Snapshot~\cite{videoavatar}, and ZJU-MoCap~\cite{neuralbody}, where we have access to human performance videos filmed by stationary monocular cameras, to show the effectiveness of our method on synthesizing novel views with high visual quality and motion dynamics.
We extensively evaluate and compare our method against existing methods, both qualitatively and quantitatively, showing that our method can exhibit state-of-the-art novel view synthesis of dynamic humans from stationary monocular cameras.
The comparisons demonstrate that directly applying existing methods that are not dedicated for our stationary monocular setting would simply lead to failure.
We also conduct ablation experiments to compare our method against several variants to understand the importance of these key designs.

\section{Related Work}

\rev{Digitizing human bodies has} received much attention in both computer graphics and computer vision, with a vast literature of human body performance capture.
We first review the works on capturing human performance.
Our work is built upon the success of the Neural Radiance Field technique, so we shall also briefly cover recent developments related to NeRF for novel view synthesis.

\noindent\textbf{Human performance capture.}
Decades of research has been devoted to faithfully capture humans performance.
Most of them follow a model-and-render procedure for producing novel view imagery.
These methods typically exploit multi-view systems~\cite{debevec2000acquiring, guo2019relightables, orts2016holoportation, collet2015high, leroy2017multi, starck2007surface}, depth cameras or fusion of depth sensors~\cite{izadi2011kinectfusion, newcombe2011kinectfusion, zhou2014color, cui2012kinectavatar, li20133d, shapiro2014rapid, zeng2013templateless, dou2016fusion4d, su2020robustfusion, dynamicfusion} to reconstruct geometry, static or motion-factored, by aggregating observation gained from various viewpoints.
Various approaches~\cite{martin2018lookingood, wu2020multi, neuralvolume, neuralbody} have also incorporated emerging neural rendering techniques for compensating the visual loss in the reconstruction.

In the sparse views case, which is more unconstrained and ambiguous, model based methods utilize the prior knowledge of parametric template models to help significantly constrain the solution space for the unobserved body parts.
The final geometry is obtained by deforming parametric coarse templates or pre-scanned accurate shapes to fit the captured images~\cite{videoavatar, carranza2003free, de2008performance, gall2009motion, stoll2010video, pons2011model, hasler2009statistical}.
Another line of works~\cite{SMPL, VIBE, hmr, spin, natsume2019siclope, saito2019pifu, saito2020pifuhd, zheng2019deephuman} learn human body priors by training networks on a large collection of images data, enabling the inference of complete human models from single images or monocular videos.
It remains, however, very difficult for these learning-based methods to produce plausible results for out-of-distribution human samples.
In general, while seeking to explicitly model the geometry and texture of the subjects, these current state-of-the-art methods have difficulty in producing realistic view synthesis by rendering from the explicitly reconstructed 3D models.

\noindent\textbf{Neural representations for view synthesis.}
Neural representation has been one of the key infrastructures to the neural rendering technique that is able to render photo-realistic imagery.
Generative Query Network (GQN)~\cite{GQN, kumar2018consistent}, the pioneering work in this direction, perceives the underlying 3D scene from a set of input images based on a neural representation and generation network. 
With an implicit notion of 3D, GQN can synthesize arbitrary views with correct occlusion.
Following that, a variety of methods~\cite{neuralvolume, srn, deepvoxels} emerged that include a more explicit representation of the 3D,  
exploiting components of the graphics pipeline.
We strongly recommend~\cite{tewari2020state} for a thorough summary of this emerging field.
Lately, the Neural Radiance Field (NeRF)~\cite{nerf} technique has revolutionized novel view synthesis of static structures by training an MLP-based radiance and opacity field.
Through a differentiable volume rendering technique, NeRF achieved unprecedented success in producing photo-realistic novel view imagery.
An explosion of NeRF techniques occurred in the research community since then that improves the NeRF in various aspects of the problem~\cite{neuralsparse, lindell2021autoint, rebain2020derf, nerfpp, nex, nerfw, nerfmm, lin2021barf, mixturevolume}.
Nevertheless, all these achievements were made on static structures, it remains challenging to extend the static NeRF to deal with dynamic scenes.

\noindent\textbf{Dynamic NeRFs.}
Lately, there has been a surge of developments related to NeRF extending to deal with dynamic scenes.
These dynamic NeRFs have shown impressive performance in view synthesis with different setups to capture the dynamics.
The most natural extension is to still use the multi-view camera setting~\cite{neuralbody, neural3d} to acquire sufficient observation of the dynamic scene.
The multi-camera setup helps constrain the modeling of the dynamics significantly.
However, these extensions require to have controlled environments and specialized hardware to acquire full observation on the dynamic scene, implying a difficulty in popularizing applications.

Another effort is to relax the need of heavy setups, by exploiting a \emph{single} free-viewpoint camera to capture dynamic scenes from \emph{varying viewpoints}~\cite{nerfies, nsff, D-NeRF, video-nerf, nonrigid_nerf, nerflow}. 
Although the extrinsic camera parameters obtained from SfM can help constrain the modeling to some extent,
the key in this monocular camera setting is to deal with the ambiguity of the geometry, texture, and motion of the subjects due to occlusion.
\rev{
The canonical NeRF and motion flow formulation in this work are similar in spirit with those recently emerging dynamic NeRFs but differ in how we regularize the challenging optimization.
}
Nerfies~\cite{nerfies} proposes an as-rigid-as-possible (ARAP) regularization of the deformation, which assumes elastic deformation behavior of the subjects. 
In addition, this ARAP regularization has extremely high computational complexity.
Analogically, NR-NeRF~\cite{nonrigid_nerf} proposes regularizers on the estimated deformations which constrain
the problem by encouraging small volume preserving deformations.
NSFF~\cite{nsff} leverages external supervision such as rough monocular depth estimation and flow-estimation, which unfortunately are not further jointly optimized during the optimization, to resolve ambiguities.
Surprisingly, although D-NeRF~\cite{D-NeRF} also uses the single free-viewpoint setting, the spirally flying camera in their setup covers the entire dynamic scene and the observation turns out to gain sufficient information.
Although these methods bypass the need for expensive equipment, they require the capture device to be empowered with certain mobility to allow capturing dynamic scenes.
NerFACE~\cite{NerFACE} shares the most similar setup to ours, but is highly specialized for human faces.
They also do not optimize to obtain face dynamics  but purely rely on the high precision of the face tracking method to capture the dynamics.

\section{Method}

\DeclarePairedDelimiter{\abs}{\lvert}{\rvert}
\DeclarePairedDelimiter{\norm}{\lVert}{\rVert}

\newcommand{\point}{ \bm{ \mathrm{x} } }
\newcommand{\viewd}{ \bm{ \mathrm{d} } }
\newcommand{\rgb}{ \bm{ \mathrm{c} } }
\newcommand{\density}{ \mathrm{\sigma} }

\newcommand{\appcode}{ \bm{ l } }
\newcommand{\im}{ \bm{I} }

\newcommand{\nerf}{ \bm{ \mathrm{F} } }
\newcommand{\mof}{ \bm{ \mathrm{M} } }
\newcommand{\moffw}{ \mof^{fw} }
\newcommand{\mofbw}{ \mof^{bw} }

\newcommand{\timestep}{ \mathrm{t} }

\newcommand{\pencoder}{ \gamma }

\newcommand{\initloss}{ \mathcal{L}^{init} }
\newcommand{\jointloss}{ \mathcal{L}^{joint} }
\newcommand{\fitmoloss}{ \mathcal{L}_{mo}^{fit} }
\newcommand{\reconloss}{ \mathcal{L}_{photo} }
\newcommand{\moreg}{ \mathcal{L}_{moco} }
\newcommand{\localmoreg}{ \mathcal{L}_{moco}^{local} }
\newcommand{\globalmoreg}{ \mathcal{L}_{moco}^{global} }

\newcommand{\freq}{ \mathrm{f} }

\label{sec:neural_rep}

\begin{figure*}[t!]
  \centering
  \includegraphics[width=0.9\textwidth]{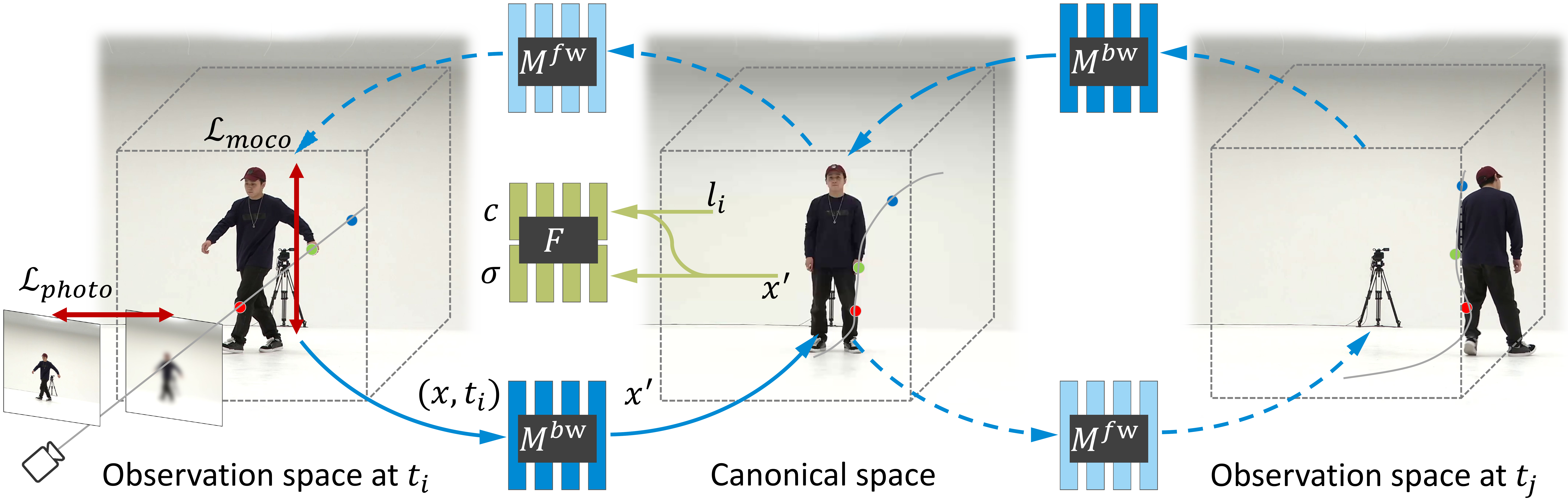}
  \caption{
  \textbf{\name}.
  The dynamic scene is represented by a shared canonical NeRF and motion flows.
  We trace rays in the observation space $\timestep_i$ and transform the samples $\point$ along the ray to 3D samples $\point'$ in the canonical space via the neural backward motion flow $\mofbw: (\point, \timestep_i) \to \point'$. 
  We evaluate the color and density of $\point$ at $\timestep_i$ through the canonical NeRF with a condition appearance code $\appcode_i$: $\nerf( \point', \appcode_{i} ) \to (\rgb, \density)$.
  The networks are initialized with rough human mesh estimation and then optimized to minimize the error $\reconloss$ of rendering captured images.
  An auxiliary neural forward motion network ($M^{fw}$) is introduced to constrain the optimization with motion consensus regularization $\moreg$ (see the loop formed by the blue arrows).
  }
  \label{fig:network}
\end{figure*}

The input to our method is a human performance video $\im=\{ \im_i \}$ captured by a stationary monocular camera, where $i \in \{0,...,m-1\}$ for $m$ observation frames.
The performer can perform arbitrary motions in front of the camera so as to show the body sufficiently for a full 360$^{\circ}$ novel view navigation.
Nevertheless,  our method can also work well for scenarios where the camera only observes the body partially.
In addition, we assume a static background image captured without the human performer.
If inapplicable, we simply set the background to white via foreground detection~\cite{maskrcnn}.

To represent dynamic scenes,
we decompose the dynamic scene that contains the moving subject into a shared canonical space represented as a neural radiance field (NeRF)
and a motion flow that models, for each time step, per-coordinate correspondences between the canonical space and the observation space  (Sec.~\ref{sec:neural_rep}). 
Both representations are \textit{simultaneously} optimized to model a dynamic scene that minimizes the error of reproducing all observation images through differentiable volume rendering.
Without the loss of generality, we set the canonical space to be the one at the first frame.

There are two key features in our work for addressing this overly ambiguous and under-constrained optimization problem.
First, we utilize domain-specific data priors 
for building up a canonical NeRF and a neural motion flow that serves as good initial values to our optimization (Sec.~\ref{sec:init}).
This still leaves a significant search space with too many irrelevant local minima.  
Second, while constraining the solution to remain close to the initial guess, we also introduce a novel motion consensus regularization. This regularization does not limit the dynamic characteristics of the moving subjects, is computationally efficient, and effectively encourages the motion to reach a high degree of consensus among all observation spaces 
(Sec.~\ref{sec:loss_reg}).

\subsection{Neural Dynamic Scenes}

\textbf{Canonical neural radiance field.}
The neural radiance field, which is approximated using an MLP network $\nerf$, is a continuous scene representation that maps a 3D coordinate $\point=(x,y,z)$ and viewing direction $\viewd$ to an emitted color $\rgb=(r,g,b)$ and volume density $\density$.
The original NeRF was introduced for synthesizing novel views of a static scene from multi-view images, where the color prediction $\rgb$ is additionally conditioned on viewing directions.
While only having constant viewing directions over training, conditioning the color on viewing directions does not hold true in our stationary monocular camera setting, this condition is thus removed from our model.
Last, similar to~\cite{nerfw}, to modulate the appearance variation across all observation images, we condition the color prediction on an optimizable appearance latent code $\appcode_{i}$ associated to each image $\im_{i}$: 
$\nerf: (\point,\appcode_i) \to (\rgb, \density)$.

\noindent\textbf{Neural motion consensus flow.}
To represent moving subjects in the dynamic scene, we introduce another MLP-parameterized network $\mofbw$ to model the motion between the canonical space and the observation space at each time step.
Formally, given a 3D coordinate $\point$ at time $\timestep_i$, $\mofbw$ is optimized to transform $\point$ \emph{back} to a 3D coordinate $\point'$ in the canonical space \rev{via predicting an SE(3) transformation~\cite{nerfies}}: 
$\mofbw: (\point, \timestep_i) \to \point'$,
where we can evaluate the color and density of $\point$ at time $\timestep_i$ with the canonical NeRF.
In addition, to enforce the motion consensus over time for penalizing the deviation of the optimization as aforementioned,
we introduce an auxiliary motion flow network $\moffw$ that inversely transforms a 3D coordinate $\point'$ in the canonical space to a 3D coordinate $\point$ in each observation space: $\moffw: (\point', \timestep_i) \to \point$.

Since directly passing raw coordinates to MLP networks would fail to learn high-frequency functions in low-dimensional problem domains~\cite{nerf}, we lift the input 3D coordinates $\point$ and the time step $\timestep$ to higher dimension spaces for both the canonical NeRF and the neural motion flow networks.
The lifting is performed using the positional encoding function $\pencoder(\point)$ and $\pencoder(\timestep)$ as proposed in~\cite{nerf}.

\noindent\noindent\textbf{Volume rendering.}
With the backward neural motion flow, we can simply evaluate the radiance field at each time step as: $\nerf( \mofbw(\point, \timestep_i), \appcode_{i} ) \to (\rgb, \density)$, for volume rendering the observation space, thus accounting for the inferred dynamics. 
The volume rendering equation as in~\cite{nerf} is employed to render images from the radiance field of each observation space.
Recall that we assume to have a decoupled static background image (i.e., without the human body). 
When rendering, the last sample on the ray is assigned with the color of the pixel corresponding to the ray on the background image.
This encourages the networks to predict high density values only for 3D coordinates of moving subjects so as to reproduce a clean background though the differentiable volume rendering.

\subsection{Optimization}

\subsubsection{Initialization}
\label{sec:init}

We use VIBE~\cite{VIBE} to estimate a SMPL~\cite{SMPL} mesh sequence from the video, with which we can sample points on the body in observation spaces and obtain their corresponding samples in the canonical space, and vice versa.
However, although VIBE has superior generalizability, it assumes an orthogonal camera projection during training over a large collection of images, 
\begin{wrapfigure}{rh}{0.15\textwidth} %
    \centering
    \includegraphics[width=0.15\textwidth]{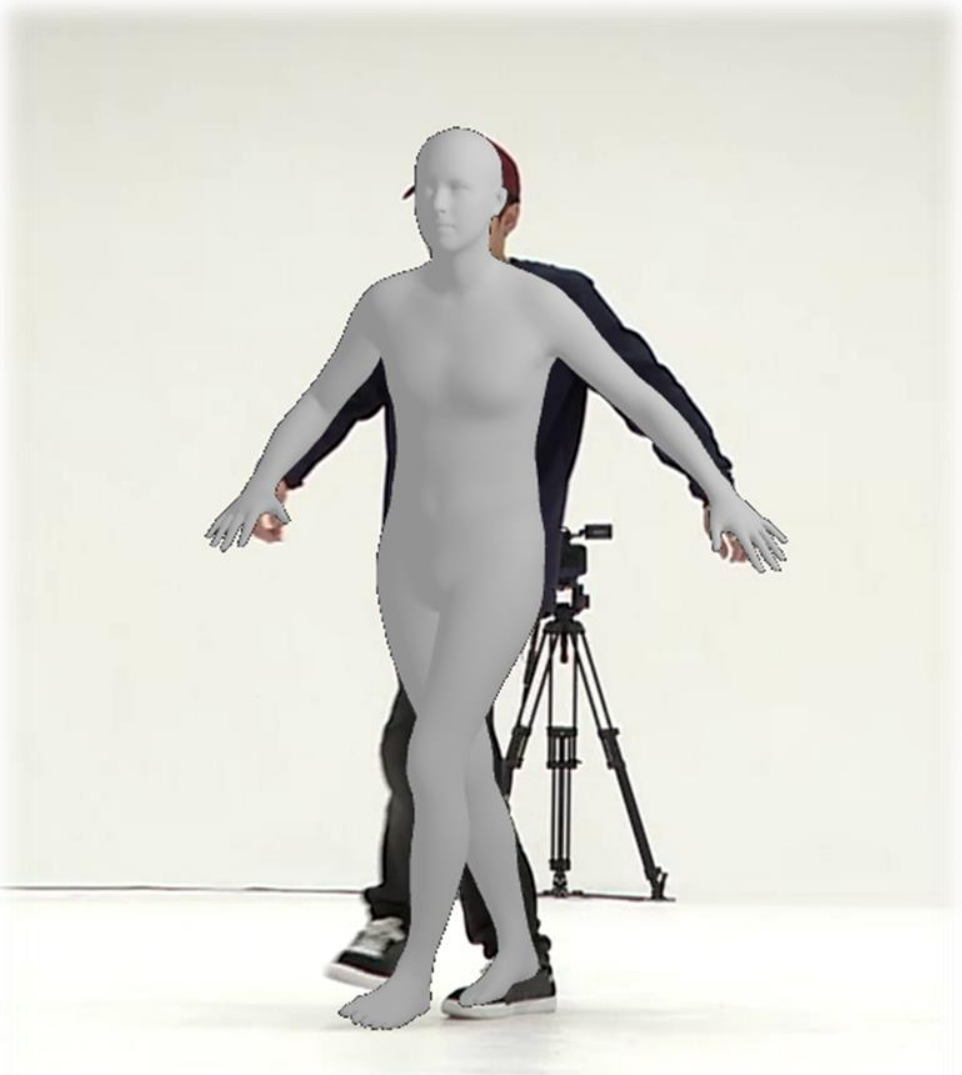}
    \caption{
    A VIBE based reconstruction provides only a rough estimate of the geometry, pose, and location.
    }
    \label{fig:mesh_estimate}
\end{wrapfigure}
which implies it cannot estimate the 3D spatial location while predicting the shape and pose parameters of the SMPL model.
Hence, to obtain the location, we render the mask of the mesh using known intrinsic camera parameters and compare against the Mask R-CNN~\cite{maskrcnn} detected mask. 
We  perform a location search on a 3D grid to minimize the matching loss.
Figure~\ref{fig:mesh_estimate} shows a resultant mesh sample.

We update the weights of both $\mofbw$ and $\moffw$ to fit the extracted observation-canonical point pairs to serve as initialization to the subsequent optimization.
Moreover, due to the infeasibility of building explicit correspondences between observation-space non-human samples and canonical-space non-human samples, we simply suppress the density value on these free samples using Binary Cross Entropy (BCE) loss without specifying the target location for them. 
Overall, the loss for initializing the motion flow networks is defined as: 
\begin{equation} 
    \begin{gathered}
        \fitmoloss := 
    \abs{ \mofbw(\point_{h}, \timestep_i) - \point'_{h} } 
  + \abs{ \moffw(\point'_{h}, \timestep_i) - \point_{h} } \\
  + \text{BCE}( \nerf_{\density}( \mofbw(\point_{f}, \timestep_i), \appcode_{i} ), 0 ),
    \end{gathered}
\end{equation}
where $\point_{h}$ denotes observation-space human samples, $\point'_{h}$ the corresponding canonical-space samples, and $\point_{f}$ free samples in observation spaces.

Further, we use the geometry of the canonical mesh to initialize the density branch $\nerf_\density$ of the canonical NeRF. 
Departing from this initialization, the neural dynamic scene represented by the canonical NeRF and the motion flow networks is subsequently optimized by the MSE photometric loss $\reconloss$ between the image rendered from the fixed viewpoint at each time step and the input image. 

\subsubsection{Regularization}
\label{sec:loss_reg}
As the initialization derived from the estimated SMPL meshes apparently contains errors regarding the geometry, location, and motion of the human (see Figure~\ref{fig:mesh_estimate}), 
the canonical NeRF and the motion flow need to be further optimized and corrected. We use two more strategies to guide the optimization process. 

\noindent\textbf{Motion consensus regularization.}
We apply a global all-to-all motion consensus regularization to the update of the motion flow networks, 
which enforces bidirectional flows between two observation spaces to achieve high consensus though the canonical space (see the loop formed by the blue solid line and dash lines in Figure~\ref{fig:network}), \\
\begin{equation}
\begin{gathered}
    \globalmoreg := \abs{ \moffw(\mofbw(\moffw(\mofbw(\point, \timestep_i), \timestep_j), \timestep_j), \timestep_i) - \point },
\end{gathered} 
\end{equation}
where $\timestep_i$ and $\timestep_j$ are random samples of time steps.
Additionally, we apply a local motion self-consensus regularization, enforcing the motion flow to be invertible locally at each time step $\timestep_i$: 
\begin{equation}
\begin{gathered}
    \localmoreg := \abs{ \moffw(\mofbw(\point, \timestep_i), \timestep_i) - \point }.
\end{gathered}
\end{equation}
The total motion consensus regularization is thus defined as: $\moreg:=\globalmoreg + \localmoreg$.
Note that we only apply this motion consensus regularization on samplings of moving subjects, while imposing no regularization on the dynamics of free space samplings.
This is achieved by simply filtering out free space samplings with a density threshold $\epsilon$=0.01.

\noindent\textbf{Coarse-to-fine flow regularization.} 
We further employ a coarse-to-fine annealing strategy, as in~\cite{nerfies}, to gradually optimize the neural dynamic scene from modeling low-frequency details to high-frequency details~\cite{tancik2020fourier}. 
This regularization is carried out by gradually increasing the frequency bands used in the positional encoding of 3D coordinates.
The positional encoding for each of the three coordinate values in $\point$ is then changed to:
$
    \pencoder(\point) = (\point, ..., w_{f-1}(\alpha)\sin(2^{\freq-1}{\pi}\point), w_{f-1}(\alpha)\cos(2^{\freq-1}{\pi}\point) ), 
$
where $\freq$ is the maximum number of frequency bands, 
$w_{k}(\alpha)=(1-\cos(\pi \text{clamp}(\alpha-k, 0, 1) ))/2$, 
and $\alpha(n) = \freq n/\mathrm{N}$ with $n$ being the current optimization iteration and $\mathrm{N}$ the total iterations for the coarse-to-fine optimization stage.
Note that this coarse-to-fine regularization is only applied to the positional encoding function of 3D locations $\pencoder(\point)$, the positional encoding of time steps $\pencoder(\timestep)$ uses constant number of frequency bands during the optimization.

\begin{figure*}[t!]
  \centering
  \includegraphics[width=\linewidth]{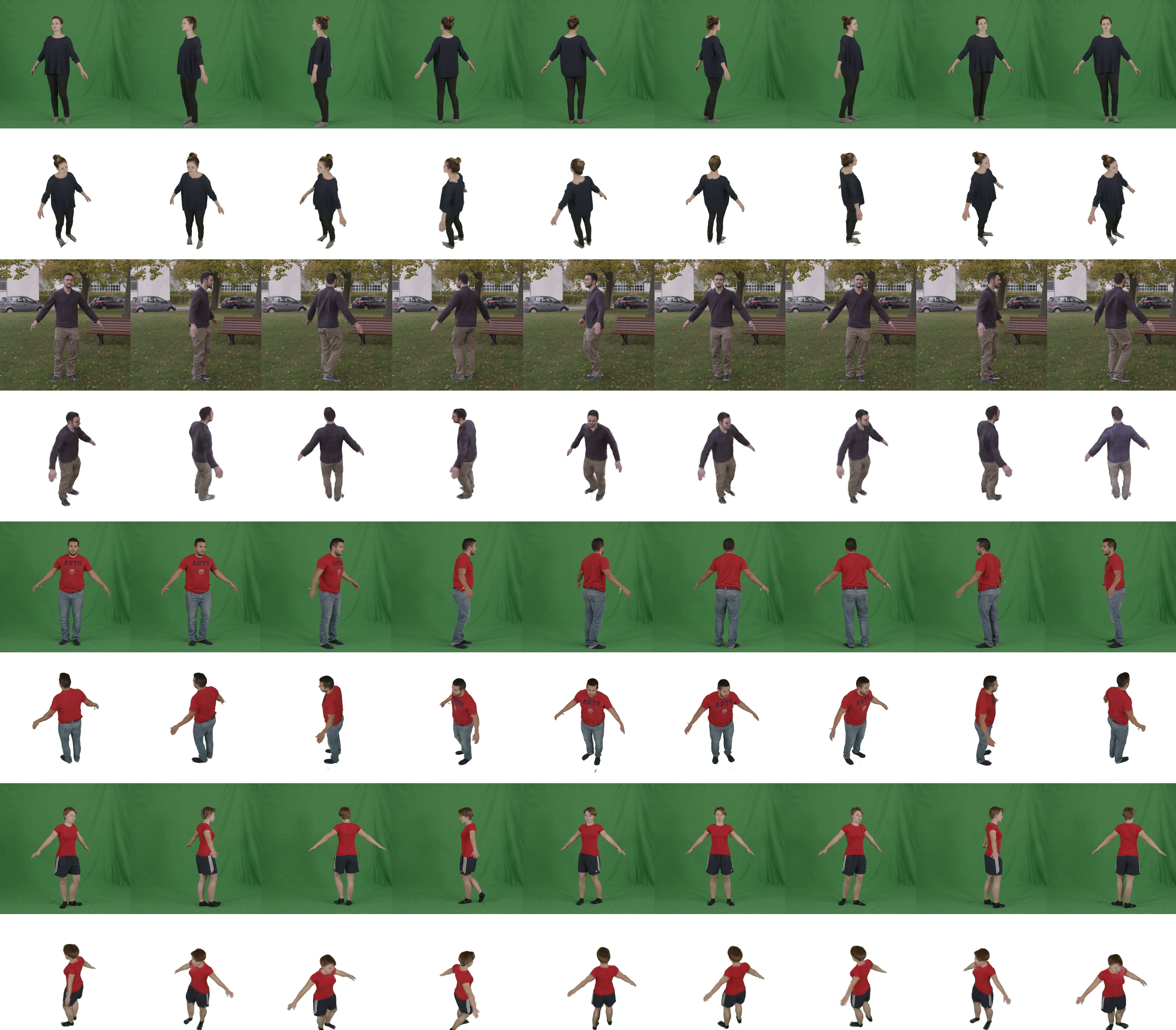}
  \caption{
    Qualitative results on People-snapshot dataset.
    Odd rows show the input observation across time from the stationary monocular view;
    even rows show the associated imagery synthesized from a novel view across time.
    Note that the logos 
    on clothing (see results in the middle of the 6th row) are recovered in the novel view, although unseen from the input view. Please use digital zoom. 
  }
  \label{fig:peoplesnapshot_results}
\end{figure*}

\subsection{Implementation details}
In our implementation, both the canonical NeRF and the motion consensus flow networks are approximated by an 8-layer MLP network with hidden width 256, ReLU activation, and a skip connection at the 4th layer.
We use hierarchical volume sampling strategy~\cite{nerf}, sampling 64 coarse locations and 128 fine locations along the rays.
We set the number of frequency bands $\freq$ used in $\pencoder(t)$ to 16,
and the maximum number of frequency bands $\freq$ used in $\pencoder(\point)$ to 8. 
We use 8 dimensions for the appearance latent codes, which are randomly initialized prior to the optimization.

\noindent\textbf{Adaptive bounding volume.}
Evaluating 3D coordinates over the whole 3D space of the dynamic scene requires the samples along the rays to be sufficiently dense for synthesizing high-quality imagery, consequently leading to a time-consuming optimization process.
Since the moving subjects typically occupy a small portion of the large space, inspired by~\cite{neuralsparse, mixturevolume}, in each observation space, we set an adaptive bounding volume for bounding the ray marching.
Concretely, at each observation space, we compute an axis-aligned bounding box (AABB) from the estimated mesh, 
and empirically enlarge this AABB by an offset of 0.2 meters along XY dimension and 0.4 meters along Z dimension to ensure it is sufficient to entirely cover the underlying human body.
Then, we only evaluate samples along the rays that are intersecting with the AABB, and bound the sampling to be within the AABB.

\noindent\textbf{Initialization and optimization.}
To initialize the density branch $\nerf_{\density}$ of the canonical NeRF, we render a set of multi-view images of the canonical SMPL mesh, and follow the optimization as in~\cite{nerf} to obtain the weights of $\nerf_{\density}$.
Next, we initialize the motion flow networks $\mofbw$ and $\moffw$ and the color branch $\nerf_{\rgb}$ of the canonical NeRF to overfit with the following loss, while freezing the density branch $\nerf_{\density}$:
$
    \initloss := \reconloss + \lambda \moreg + \mu \fitmoloss,
$
where we use $\lambda=0.2$, and $\mu=10$.
Subsequently, we unfreeze $\nerf_{\density}$ and jointly optimize all networks with the loss: 
$
    \jointloss := \reconloss + \lambda \moreg.
$
Typically, the entire optimization takes around 3 days on 8 V100 GPUs.
Rendering a novel view image at resolution 512 x 512 takes roughly 30 seconds.
Since it is ambiguous to model the background from stationary monocular cameras, we simply set a background image with solid colors (a virtual background image is also possible) for rendering novel views.
See also the supplementary.

\section{Experiments}

\begin{figure*}[]
  \centering
  \includegraphics[width=\linewidth]{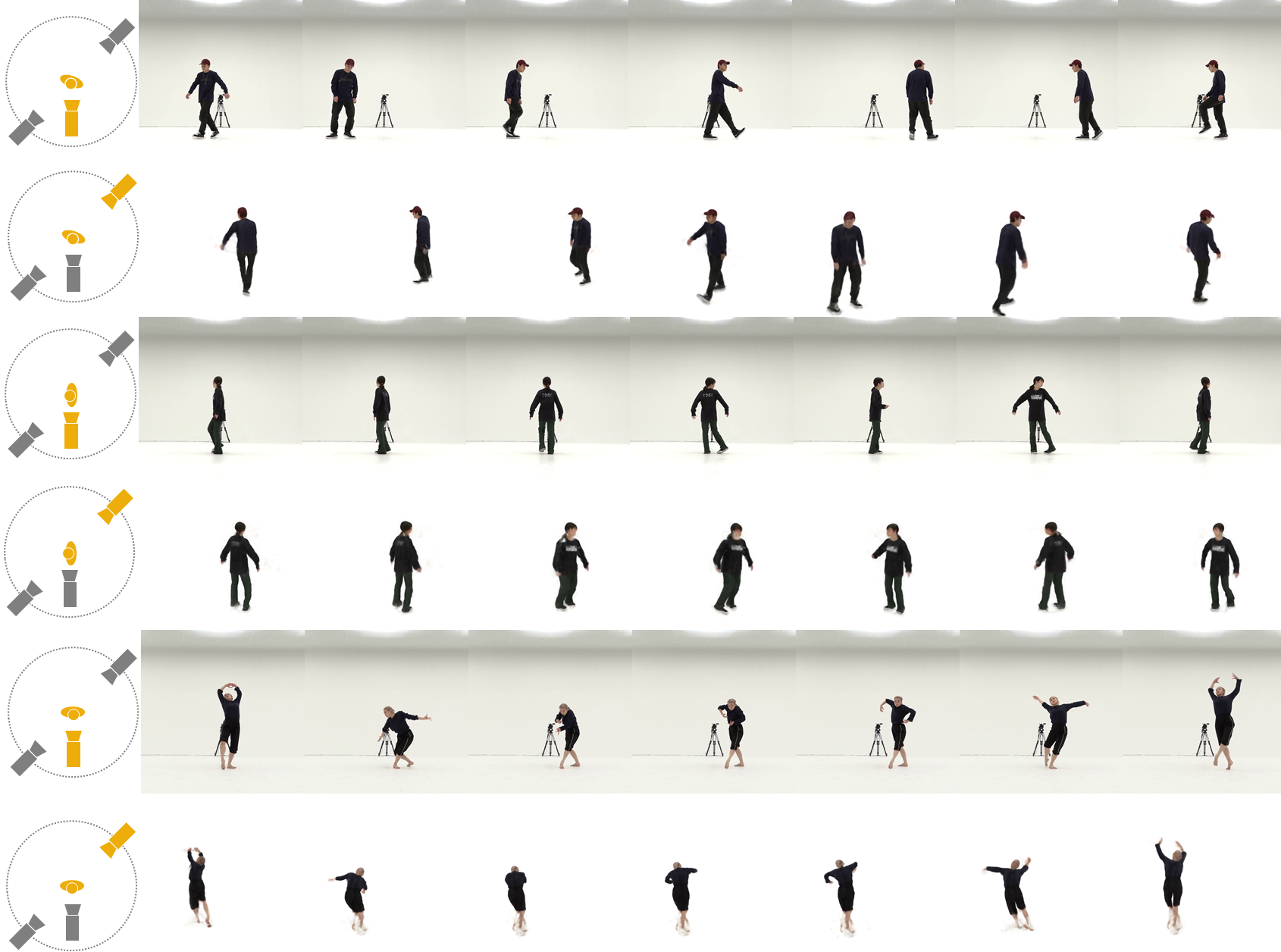}
  \caption{
    \rev{
    Qualitative results on AIST. From top to bottom: male pop, female pop, and female ballet dancer.
    Each two rows shows the input observation across time (the row showing gray background) and associated imagery from a novel view.
    The first column visualizes the camera setup and the initial body orientation in the first frame.
    }
  }
  \label{fig:aist_results}
\end{figure*}

\begin{figure}[]
  \centering
  \includegraphics[width=0.9\linewidth]{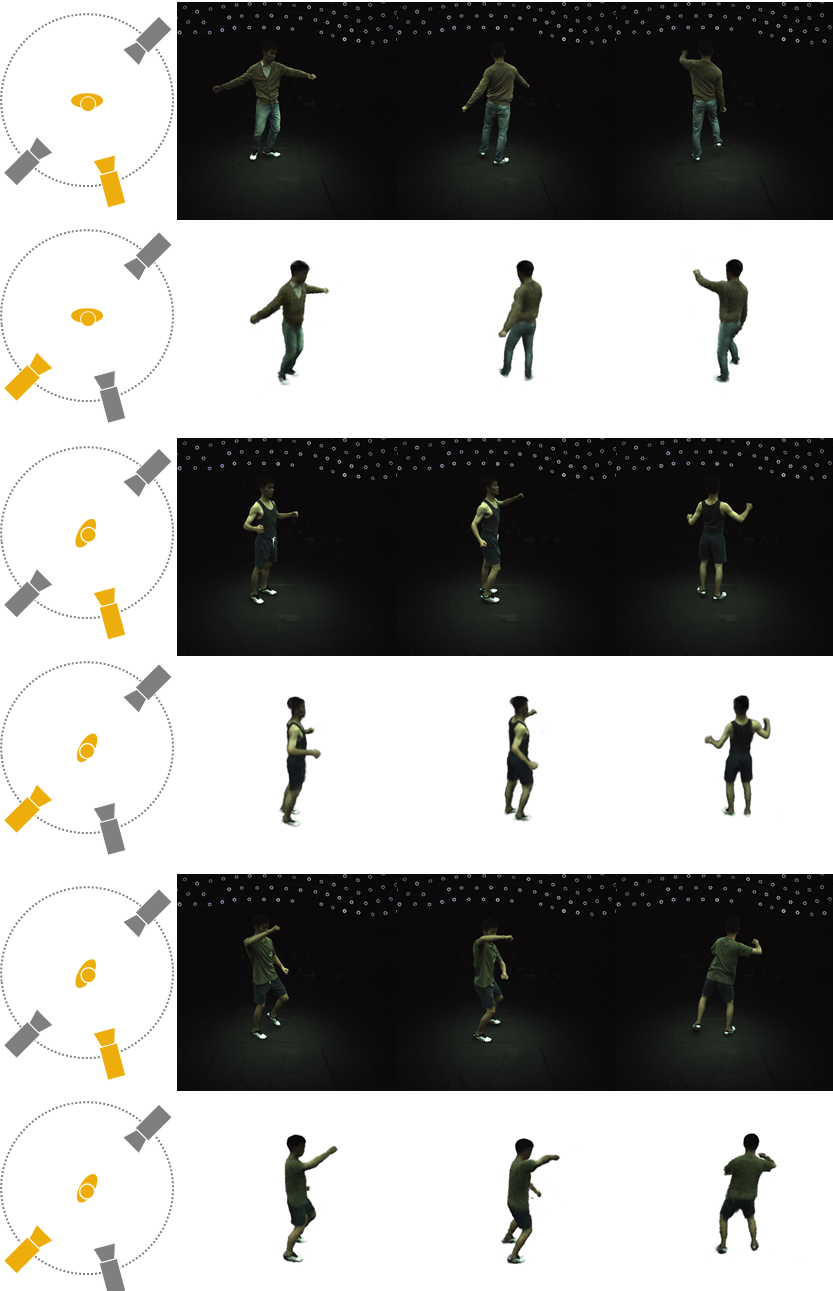}
  \caption{
    \rev{Qualitative results on ZJU-MoCap.
    From top to bottom: Swing2, Warmup, and Swing1.
    Each two rows shows the input observation across time (the row showing garyish background) and associated imagery from a novel view.
    The first column visualizes the camera setup and the initial body orientation in the first frame.}
  }
  \label{fig:aist_results}
\end{figure}

In this section, we describe the datasets and metrics for evaluating our method, and show both qualitative and quantitative results of our method.
We also present quantitative and qualitative comparisons against
several baseline methods, along with experiments conducted for evaluating different aspects of our method.

\noindent\textbf{Datasets.}
We evaluate on three data sources that contain human performances of varying complexity:
\textit{(A) People-snapshot}~\cite{videoavatar}, which captures human performers that rotate while holding an A-pose in front of a stationary camera.
This dataset contains motions of rather low complexity, and does not offer ground truth images from other viewpoints.
So we further evaluate on more challenging datasets, where ground truth novel view images are available as well.
\textit{(B) AIST}~\cite{aist, aistpp}, which is a shared database containing dance videos.
The videos are captured using multiple cameras (9 at most) surrounding a dancer to simultaneously shoot from various directions.
We use only the monocular video captured from the lower front by default (i.e., the camera with ID C09 as described in the database) for modeling, and use videos filmed from the rest positions as ground truth for evaluation.
\textit{(C) ZJU-MoCap}~\cite{neuralbody}, which is another human performance dataset created for evaluating dynamic human reconstruction from multi-view videos.
The humans perform arbitrary and complex motions, including twirling, arm swings, punching, kicking and so forth, in a multi-camera system that has 21 synchronized cameras.
Again, we use only the monocular video captured at the camera with ID 01 by default, and use the remaining cameras for evaluation.
We refer readers to the supplementary for more details of data processing.

\noindent\textbf{Evaluation measures.}
Since it is ambiguous to infer the surrounding environment in novel views given a stationary monocular camera only, we simply mask out the background of ground truth images and crop an image patch (512 x 512) around the human center for computing the metrics.

There are several commonly  standard metrics: peak signal-to-noise ratio (PSNR), and perceptual similarity through LPIPS~\cite{zhang2018unreasonable}.
However, we found that these metrics are very unsuitable for our task due to severe misalignment between the ground truth images and the synthesized views (which we shall discuss in the limitations), exhibiting an irrational trend. This is also pointed out in~\cite{nerfies}.
Here we first conduct experiments to investigate the influence of the misalignment on the PSNR and LPIPS metrics.
Concretely, we generate a set of images via translating the ground truth image by offsets (in pixel) along random directions or rotating around the human center by degrees along random directions.
Table~\ref{tab:effect_misalignment} shows the PSNR and LPIPS scores degrade dramatically while the visual content remains correct but misalignment increases.

So, to better evaluate our novel view synthesis results, 
we propose to calculate the \emph{plausibility} of the synthesized images as human imagery, which is evaluated as the human detection accuracy in percentage produced by Mask-RCNN. 
In addition, we further propose to measure the \emph{pose accuracy} that is evaluated as the commonly used object keypoints similarity (OKS) in the human pose detection field.
Concretely, we use the AlphaPose~\cite{fang2017rmpe, li2018crowdpose, xiu2018poseflow} method to detect the human keypoints both in the ground truth image and the synthesized novel view image, 
then canonicalize these two detected human poses by aligning their mean centers,
and finally compute the OKS as the pose accuracy for the novel view image.

\begin{table}[b]
  \centering
  \small
  \caption{PSNR and LPIPS scores degrade dramatically while the misalignment increases but the visual content remains correct.}
    \begin{tabular}{lcccccc}
\cmidrule{1-6}    translation (px) & 10    & 20    & 30    & 40    & 50    &  \\
    PSNR  & 20.46 & 17.44 & 16.38 & 15.51 & 14.84 &  \\
    LPIPS & 0.04  & 0.08  & 0.10  & 0.12  & 0.13  &  \\
\cmidrule{1-6}          &       &       &       &       &       &  \\
    \midrule
    rotation (deg.) & 5     & 10    & 15    & 20    & 25    & 30 \\
    PSNR  & 21.07 & 18.24 & 16.97 & 16.25 & 15.74 & 15.34 \\
    LPIPS & 0.03  & 0.06  & 0.08  & 0.10  & 0.11  & 0.12 \\
    \bottomrule
    \end{tabular}%
  \label{tab:effect_misalignment}%
\end{table}%

\begin{table*}[t]
    \centering
    \small
    \caption{Quantitative comparisons on AIST dataset. D-NeRF outputs blank imagery at novel views.
    }
    
    \resizebox{\textwidth}{!}{
    
    \setlength\tabcolsep{2pt} %
    
    \begin{tabular}{l|ccccc|ccccc|ccccc|ccccc}
    \toprule
          & \multicolumn{5}{c|}{PSNR}             & \multicolumn{5}{c|}{LPIPS}            & \multicolumn{5}{c|}{\textbf{Plausibility} $\uparrow$} & \multicolumn{5}{c}{\textbf{OKS} $\uparrow$} \\
\cmidrule{2-21}          & D-NeRF & NSFF  & NB    & NerFACE & MoCo-Flow & D-NeRF & NSFF  & NB    & NerFACE & MoCo-Flow & D-NeRF & NSFF  & NB    & NerFACE & MoCo-Flow & D-NeRF & NSFF  & NB    & NerFACE & MoCo-Flow \\
    \midrule
    male pop & 15.575 & 13.514 & \textbf{18.195} & 17.008 & 16.102 & 0.133 & 0.561 & \textbf{0.094} & 0.104 & 0.109 & 0.000  & 0.060  & 0.430  & 0.640  & \textbf{0.941 } & 0.000  & 0.092  & 0.288  & 0.431  & \textbf{0.670 } \\
    female pop & 16.578 & 15.131 & \textbf{17.694} & 16.425 & 14.531 & 0.117 & 0.511 & \textbf{0.086} & 0.099 & 0.151 & 0.000  & 0.020  & 0.320  & 0.563  & \textbf{0.943 } & 0.000  & 0.079  & 0.337  & 0.383  & \textbf{0.495 } \\
    female ballet & 16.912 & \textbf{17.890} & 17.880 & 17.215 & 17.265 & 0.123 & 0.104 & 0.104 & 0.118 & \textbf{0.100} & 0.000  & 0.030  & 0.360  & 0.605  & \textbf{0.920 } & 0.000  & 0.018  & 0.134  & 0.113  & \textbf{0.503 } \\
    \bottomrule
    \end{tabular}%
    
    }
    
    \label{tab:comparisons}
\end{table*}

\begin{figure}[t!]
    \centering
    \includegraphics[width=\linewidth]{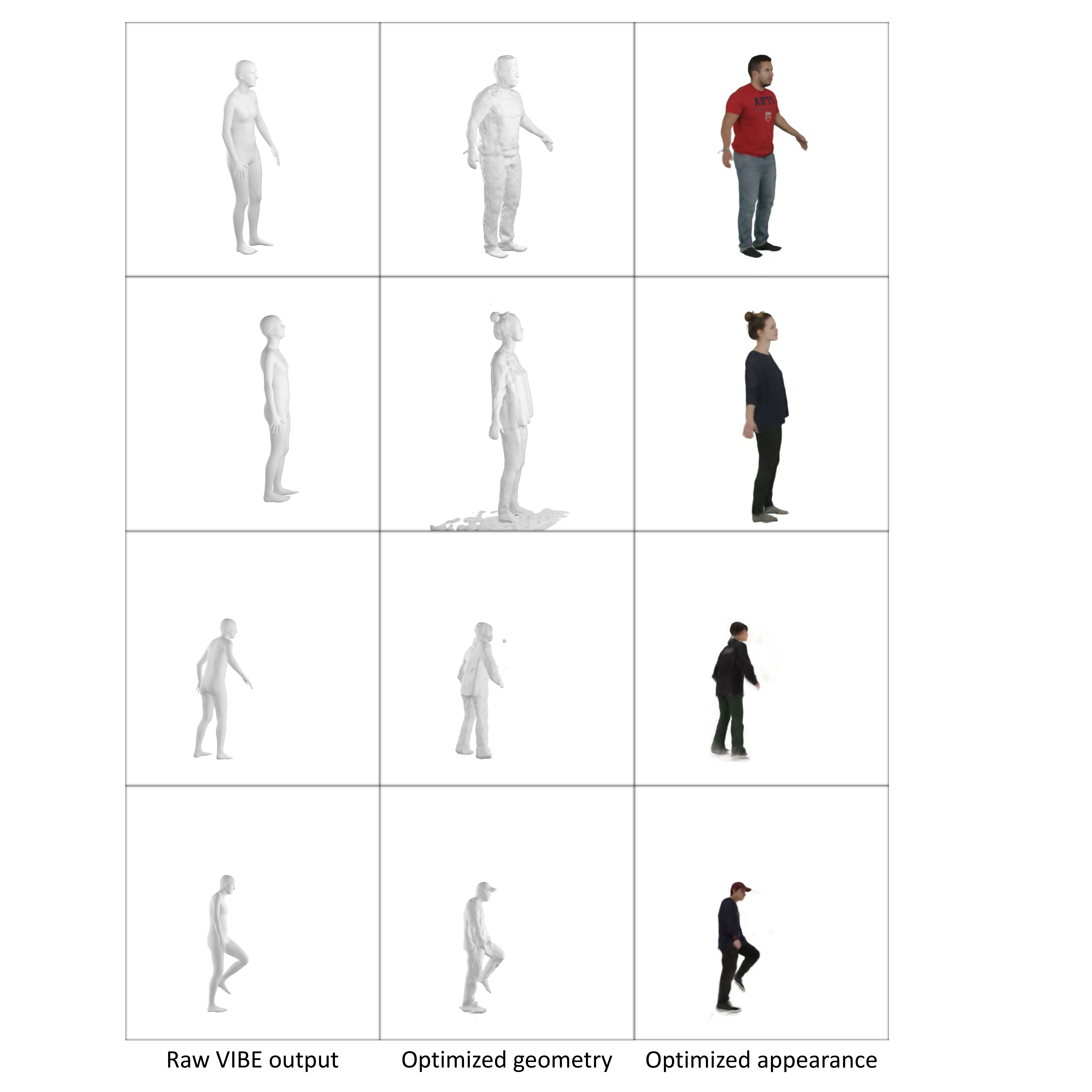}
    \caption{
    Side-by-side visual comparisons to raw VIBE outputs, showing the improvements on both the geometry and the appearance over the optimization.
    }
    \label{fig:raw_vibe_comparison}
\end{figure}

\begin{figure}[b]
    \centering
    \includegraphics[width=\linewidth]{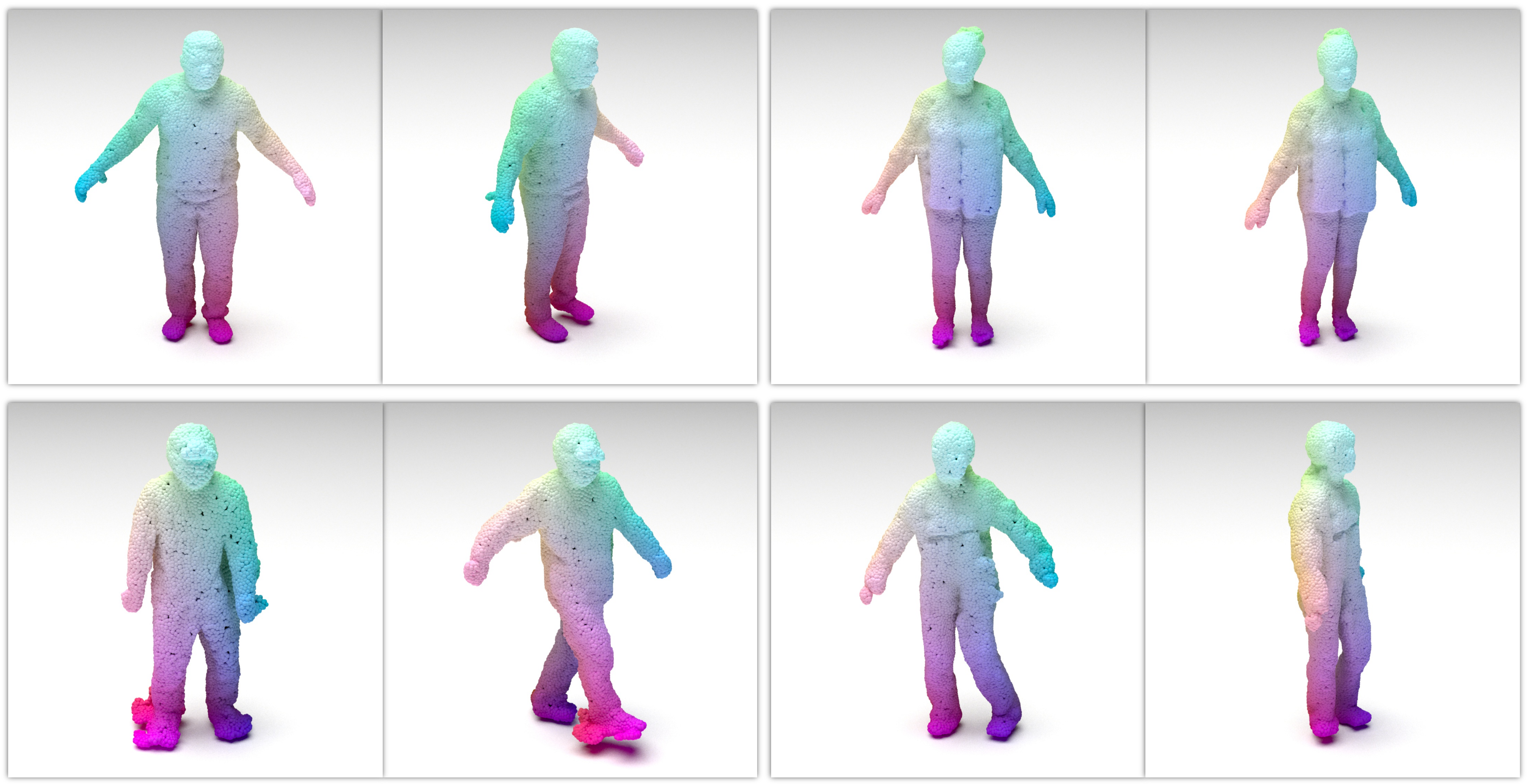}
    \caption{
    Each pair visualizes the dense correspondences derived from the backward motion flow between the canonical body (left) and the body reconstructed (right) at an observation frame.
    }
    \label{fig:flow_visualizatoin}
\end{figure}

\noindent\textbf{\name novel view synthesis.}
We present qualitative results of \name on synthesizing novel views of dynamic humans given a stationary monocular video only.
Figure~\ref{fig:peoplesnapshot_results} and Figure~\ref{fig:aist_results} show the visual results on the aforementioned datasets, wherein we render the imagery from \emph{unseen} viewpoints in the dynamic scene during the performance.
We can see, in addition to reconstructing highly plausible motions and human geometries with clothing in novel views, our method can also recover fine details such as patterns on clothing, the thin string-band on the wrist, and the hat brim in the novel view imagery.

In addition, we show more qualitative evaluation of the learned motion flow.
In Figure~\ref{fig:raw_vibe_comparison}, we present side-by-side visual comparisons of \name results against the raw VIBE outputs, showing the improvements of the optimized geometry and appearance by the motion flow over the initialization.
In Figure~\ref{fig:flow_visualizatoin}, we visualize the dense correspondences derived from the learned motion flow between the canonical and the observation space.

\noindent\textbf{Comparisons.}
We present both qualitative and quantitative comparisons against several latest works, namely D-NeRF~\cite{D-NeRF}, NSFF~\cite{nsff}, Neural Body (NB)~\cite{neuralbody} \rev{, and NerFACE~\cite{NerFACE}}, on the AIST dataset, where ground truth novel views are available for evaluation.
\rev{
We obtained the results of the former three using authors' code.
As for NerFACE, which is highly specialized for faces and does not optimize the dynamics as aforementioned, we implement it based on our framework to work on SMPLs. 
More details can be found in the appendix.
}

\begin{figure}[b]
    \centering
    \includegraphics[width=0.9\linewidth]{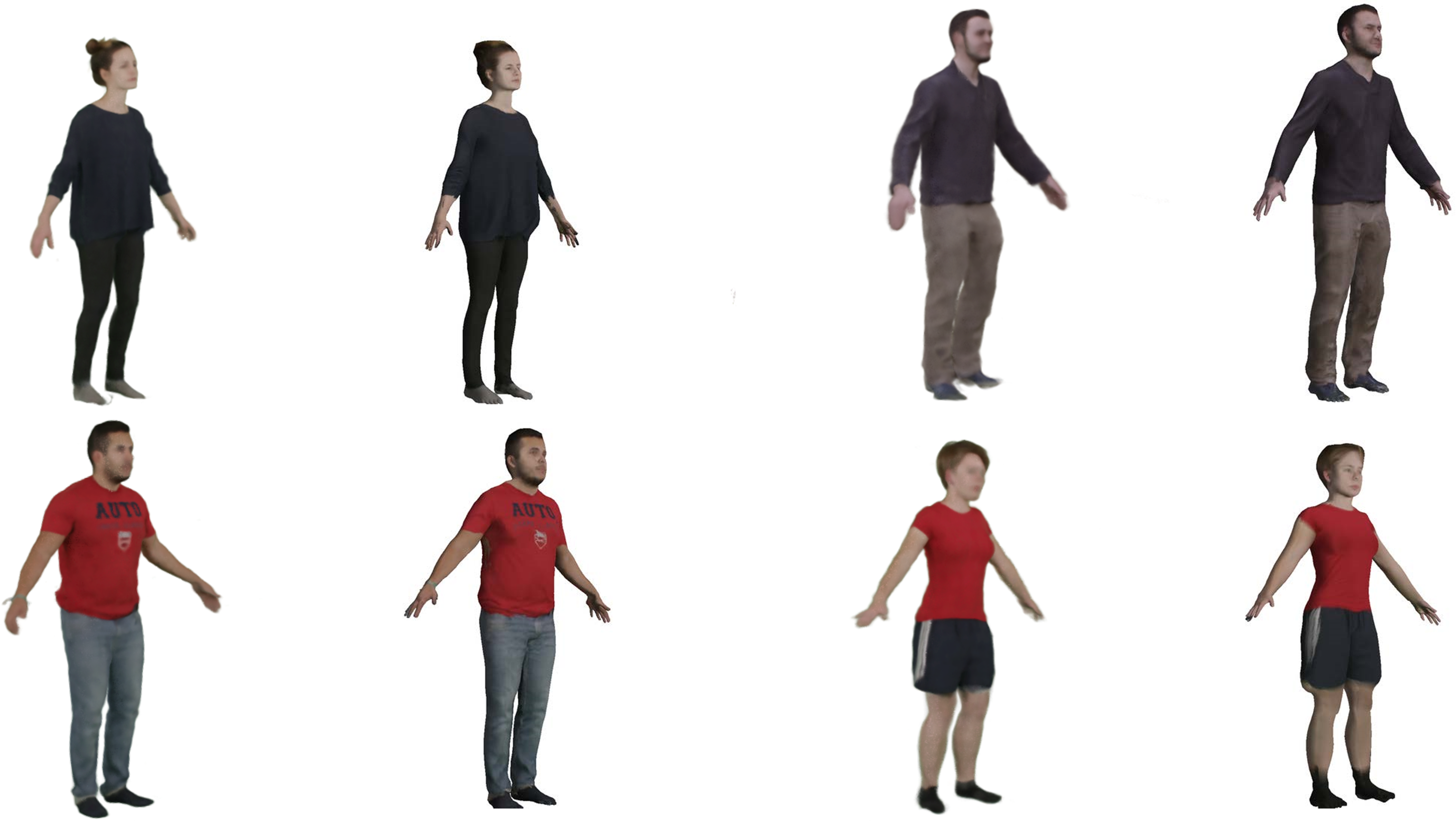}
    \caption{
    \rev{Qualitative comparison with Video Avatar. Each pair shows the visual result of ours (left) and Video Avatar (right).}
    }
    \label{fig:video_avatar_comparison}
\end{figure}

\begin{figure*}[]
    \centering
    \includegraphics[width=0.8\linewidth]{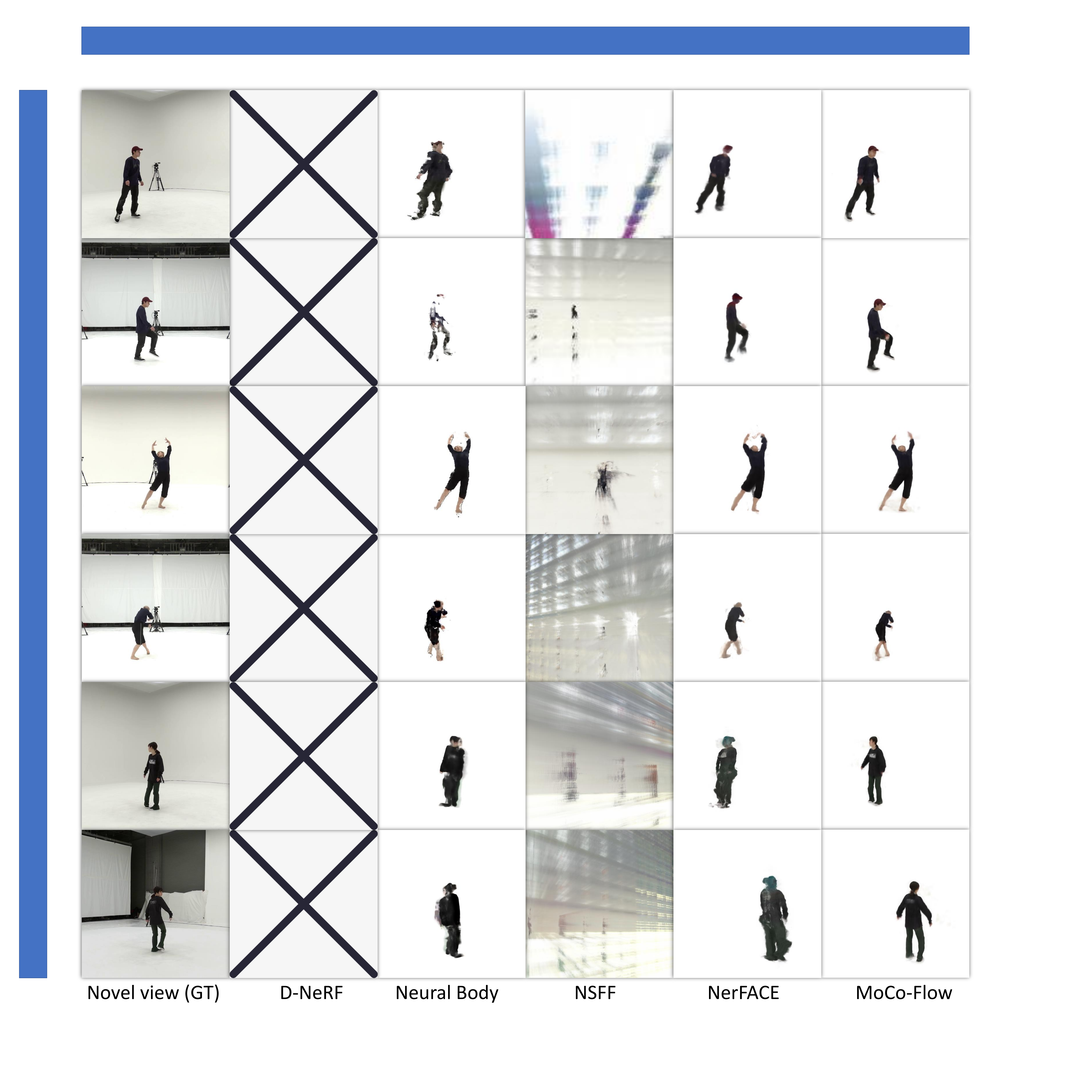}
    \caption{Side-by-side visual comparisons.
    D-NeRF failed on the task and outputs blank imagery.
    Our method has the best visual quality and plausibility.
    }
    \label{fig:visual_sidebyside}
\end{figure*}

Although these methods can still overfit the training view,
directly applying existing methods that are not dedicated for our stationary monocular setting would simply lead to failure in synthesizing novel views.
D-NeRF even outputs blank imagery.
The quantitative comparisons are presented in Table~\ref{tab:comparisons}.
We struggle to obtain higher PSNR and LPIPS due to the \emph{mismatch} between the physical factors of the reconstructed 3D and the ground truth, as PSNR and LPIPS are not ideal metrics for evaluation on our task,
while the baselines easily get similar scores with meaningless results (e.g., the D-NeRF even gets high scores with white images).
But, our method dominates over the plausibility and pose accuracy score, outperforming the baselines by significant margins, as baselines completely failed to synthesize human imagery from novel views.
We present the visual comparisons in Figure~\ref{fig:visual_sidebyside}, where our results have high plausibility and accurate poses, and outperform baselines in consistence with the quantitative comparisons.

\rev{Last, we also show side-by-side comparisons with a model-based method - Video Avatar~\cite{videoavatar}, which works only on A-posed performers and reconstructs only the canonical textured mesh.
The visual comparison results are presented in Figure~\ref{fig:video_avatar_comparison}.
}

\begin{figure}[]
  \centering
  \includegraphics[width=\linewidth]{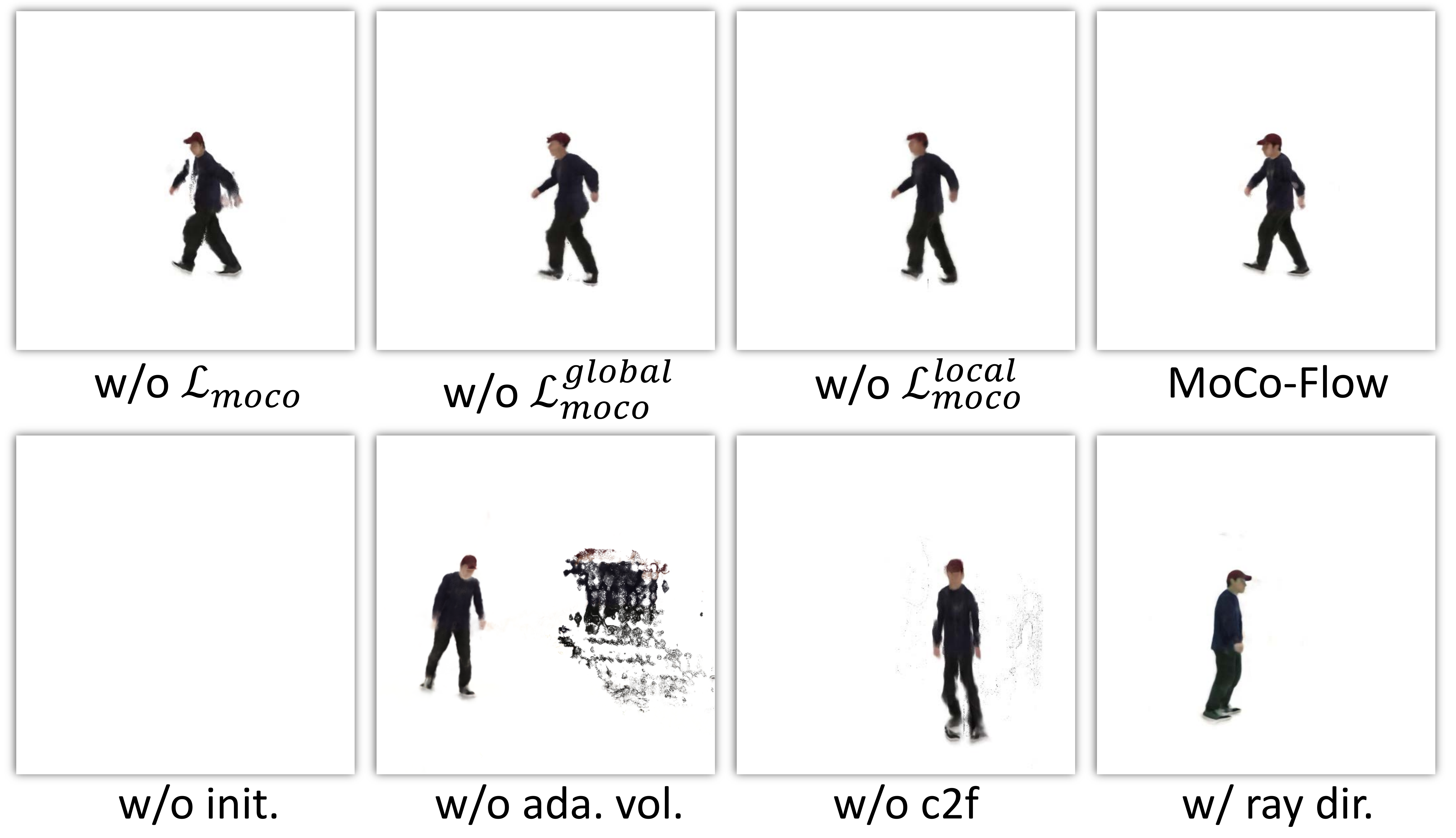}
  \caption{
    \rev{Qualitative results of removing each component from \name.}
  }
  \label{fig:ablation}
\end{figure}

\noindent\textbf{Ablation study.}
\rev{
We first conduct experiments, where several variants are obtained by removing one component from \name, to individually evaluate the contribution of each component.
}
Although most variants are still able to overfit the input observation, 
they fail to produce high-quality novel view images.
More specifically,
without the total (\textit{w/o $\moreg$}), the global (\textit{w/o $\globalmoreg$}), or the local motion consensus regularization (\textit{w/o $\localmoreg$}), the body are significantly more distorted;
\begin{wrapfigure}{r}{0.15\textwidth} %
    \centering
    \includegraphics[width=0.15\textwidth]{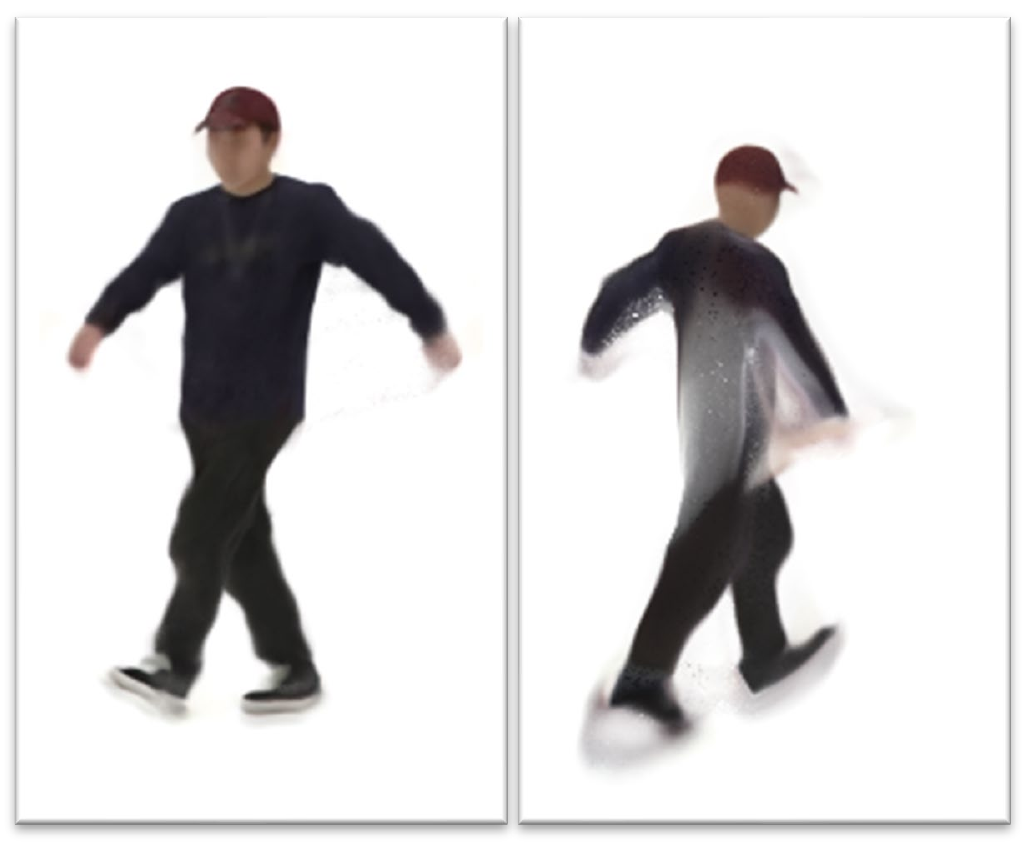}
    \caption{\rev{Effect of partial observation during capture.}}
    \label{fig:partial_ob}
\end{wrapfigure}
without the dedicated initialization step (\textit{w/o init.}), the optimization completely failed, producing meaningless blank novel view imagery;
with the absence of the adaptive bounding volume (\textit{w/o ada. vol.}), we observe plenty of noisy floaters in novel views;
without the coarse-to-fine regularization (\textit{w/o c2f}), the optimization is very unstable and leads to noisy imagery;
with the appearance branch conditioning on the ray direction (\textit{w/ ray dir.}), the results exhibit abnormal colors under novel views.
The quantitative and qualitative results are presented in Table~\ref{tab:ablation} and Figure~\ref{fig:ablation}, respectively.
\rev{
Moreover, to offer a more intuitive understanding from another perspective, we conducted additional ablation experiments, wherein variants are created by progressively adding modules to a base model that purely consists of a canonical NeRF and motion networks. The quantitative results are presented in Table~\ref{tab:ablation_seq}. Note the significant gain achieved when introducing our initialization with SMPL, motion consensus regularization, and adaptive volume.
}

\begin{table*}[h!]
  \centering
  \small
    \caption{\rev{Quantitative results of removing each component from \name .} }
    
    \begin{tabular}{l|rrrrrrrr}
    \toprule
          & \multicolumn{1}{l}{w/o $\moreg$} & \multicolumn{1}{l}{w/o $\globalmoreg$} & \multicolumn{1}{l}{w/o $\localmoreg$} & \multicolumn{1}{l}{w/o init.} & \multicolumn{1}{l}{w/o ada. vol.} & \multicolumn{1}{l}{w/o c2f} & \multicolumn{1}{l}{w/ ray dir.} & \multicolumn{1}{l}{MoCo-Flow} \\
    \midrule
    Plausibility $\uparrow$ & 0.659  & 0.865  & 0.851  & 0.000  & 0.826  & 0.726  & 0.940  & \textbf{0.941 } \\
    OKS $\uparrow$   & 0.609  & 0.641  & 0.634  & 0.000  & 0.579  & 0.512  & 0.663  & \textbf{0.670 } \\
    \bottomrule
    \end{tabular}%
    
  \label{tab:ablation}%
\end{table*}%

We also investigate the performance of our method under the partial capture situation, 
wherein the camera does not fully capture the body.
To this end, we simply cut out the forepart from the video of the male pop dancer and stop at where the dancer is about to take the spinning movement and show his back.
We observed that our method failed to infer and complete the imagery of the missing regions on the back, producing erroneous colors due to the lack of observation on the back (see the right of the inset Figure~\ref{fig:partial_ob}).
Nevertheless, our method is still able to produce correct imagery from the views near the front view of the dancer (see the left of the inset Figure~\ref{fig:partial_ob}).

\section{Conclusion}

We have presented a dynamic NeRF technique for synthesizing novel views of dynamic humans from stationary monocular cameras.
Without the observation from various viewpoints to constrain the dynamics modeling, the problem is overly unconstrained and ambiguous. We address this problem with a carefully designed 
optimization scheme, which disambiguates bad local minima early at the initialization phase,
and imposes a crucial regularization on the motion flow update to reach a high degree of \emph{consensus} across the observations. This regularization has been demonstrated to be effective on preventing the optimization from deviating too much from the initialization and from landing on bad local minima.

\textit{Limitations and future work.}
There are, nevertheless,  limitations to our method in its current form.
As with any monocular method, physical scale remains inherently ambiguous, thus, it is not guaranteed that our method can model the human body with correct physical scale (e.g., we observed that the modeled human has longer legs when the estimated mesh in VIBE output leans forward). 
Generally speaking, our method inherits erroneous estimation of the human body (VIBE's output in our case), and cannot compensate large errors in pose, location, and geometry of the human body.
Moreover, the motion of self-occluded parts may not be correctly modeled, due to vanishing gradients from the photometric reconstruction loss.
Nevertheless, with the rapid advances in learning human priors, we believe that the above issues will be  alleviated by stronger priors.
It would also be an interesting direction to explore how the proposed method can be extended to general objects if given proper priors.
Lastly, as we represent the dynamics using per-coordinate dense motion flows, evaluating and optimizing the networks are computational intensive and time-consuming.
In the future, we would like to explore the possibility of leveraging a hybrid model that brings together the explicit parametric model and neural-based implicit model, possibly using local implicits, to greatly reduce the optimization space and computation time.
This would also be helpful for extending the proposed method to much longer video clips.

\begin{table}[h]
  \centering
  \small
  \caption{\rev{Quantitative results of progressively adding modules to a base model consisting of a canonical NeRF and motion networks.}}
    \begin{tabular}{l|cc}
    \toprule
          & Plausibility & OKS \\
    \midrule
    Base  & 0.000  & 0.000  \\
    Base + c2f & 0.000  & 0.000  \\
    Base + c2f + init. & 0.618  & 0.486  \\
    Base + c2f + init. + moco & 0.826  & 0.579  \\
    Base + c2f + init. + moco + ada. vol. & 0.941  & 0.670  \\
    \bottomrule
    \end{tabular}%
  \label{tab:ablation_seq}%
\end{table}%

\rev{
\section*{Acknowledgements}
We thank the anonymous reviewers for their insightful comments and feedback. This work is supported in part by grants from 
the Joint NSFC-ISF Research Grant (62161146002), and gifts from Adobe Research.
}

\bibliographystyle{eg-alpha}
\bibliography{ref}

\clearpage

\appendix

\section{Effect of the initialization errors}
\begin{figure}[h]
  \centering
  \includegraphics[width=\linewidth]{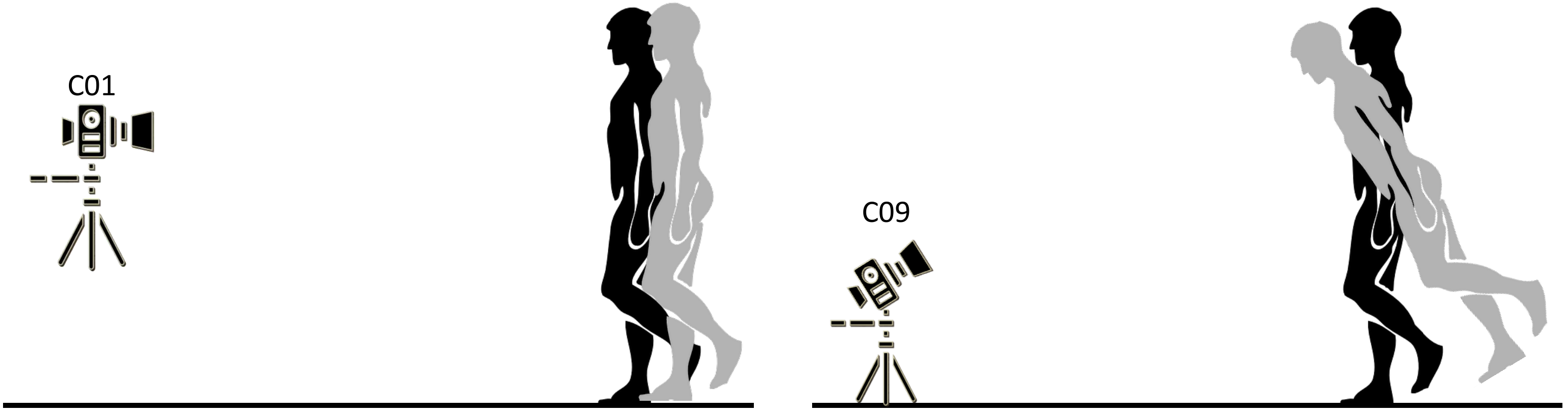}
  \caption{
    Left: with the camera C01 in AIST, which is horizontally placed, the pose of the VIBE estimated mesh (in light gray color) is more accurate. But note that there is still a considerable amount of translation error; right: with the camera C09, which has a upward view, the pose of the VIBE estimated mesh has more rotation errors to the ground truth (in aterrimus).
  }
  \label{fig:09_vs_01}
\end{figure}

\begin{figure}[]
  \centering
  \includegraphics[width=\linewidth]{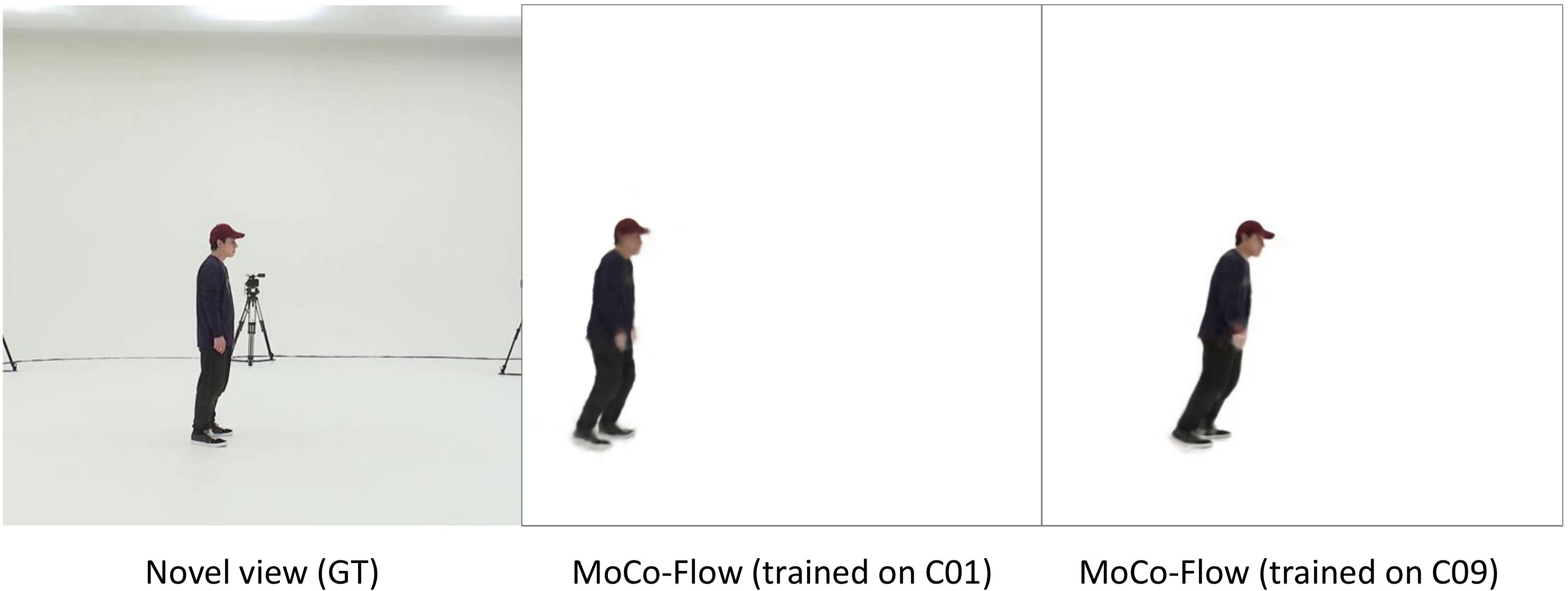}
  \caption{
    Left: the ground truth view from the side;
    middle: training on AIST-C01 data produces imagery with more accurate human poses but larger translation errors;
    right: training on AIST-C09 data leads to imagery wherein the human body leans more forward.
  }
  \label{fig:09_vs_01_results}
\end{figure}

As discussed in the limitations, the misalignment occurs in our results as our method inherits erroneous estimation of the human body, and cannot compensate large errors in  pose, location, and geometry of the human body.
Moreover, we found that the VIBE tends to estimate human poses that are parallel to the image projection plane, which implies that the pose estimation is less accurate when the camera is not placed horizontally, thus the derived initialization leads to worse local minima (see Figure~\ref{fig:09_vs_01}).
So, we also show the results of our algorithm trained on AIST camera C01 (AIST-C01 in short) data to demonstrate the performance of our method with different camera viewpoint and initialization (see Figure~\ref{fig:09_vs_01_results}).

\section{Additional implementation details}

\paragraph{Optimization details}
We optimize the networks using the Adam optimizer with a learning rate linearly decayed by a factor of 0.9999 until the maximum number of iterations is reached.
We sample 384 rays on a randomly selected image and sample 192 points (64 at the coarse level and 128 at the fine level) along each ray for each iteration of the optimization.
The initialization takes $N_1$ iterations to converge, 
followed by $N_2$ iterations for the coarse-to-fine joint optimization (i.e., linearly anneal $\alpha$ from 0 to 8 over the iterations).
We further keep $\alpha$ at 8 for $N_3$ iterations to fine-tune for more high-frequency details.
More specifically, we set $(N_1, N_2, N_3) = (200K, 1500K, 1000K)$, $(N_1, N_2, N_3) = (500K, 1500K, 1000K)$, and $(N_1, N_2, N_3) = (800K, 2000K, 1500K)$ for People-snapshot, AIST, and ZJU-MoCap, respectively.

\paragraph{Test}
As discussed in the limitations, our method inherits the erroneous estimation of the human body from the VIBE output.
Typically, we found that, on AIST dataset, VIBE often estimates human bodies with large pose and location errors.
So, in order to synthesize more visually pleasing results on AIST camera C09 data (AIST-C09, in short), 
we further introduce a post-processing to manually adjust the orientation of the reconstructed dynamic scenes to be roughly upright.
Note that the misalignment to the GT would still exist even with this rough re-orientation post-processing.

\rev{
\paragraph{Baseline details}
The results of D-NeRF, Neural Body, and NSFF are obtained with their released code.
As for NerFACE, since it is highly specialized for faces and takes as input the face parameters, we implement it in our framework to work on SMPLs.
More concretely, 
instead of rigidly transforming the whole observation-space volume with the estimated face parameter, 
we convert only the points that are near to the SMPL estimate within a distance threshold (0.2m in our implementation) using the transformation matrix of its nearest vertex on SMPL.
Then, as described in their paper, the transformed point, a learnable code, and the corresponding SMPL parameters are fed into the neural radiance field for training.
NSFF only supports reconstruction in NDC space, it is non-trivial to adapt it to work on non-NDC space, which is also mentioned in their official code repository.
Although the novel view test results of NSFF are obtained in an approximate way where the learned NDC cubic space is scaled to fit the target physical volume,
we highlight that it is the inability to model the target volume from a single stationary view that accounts for the improper content that is revealed to be unmeaning at novel views.

Last, it is rather easy to configure the cameras, for all methods, in our setting. 
All cameras are stationary and hence are set to be aligned with a world coordinate system. 
Near and far planes are set correctly in baselines following their description and instructions, so samplings are concentrated in a proper volume. 
}

\section{Datasets}
We present more details of data processing:
\textit{(A) People-snapshot}:
each video in this dataset lasts around 10 seconds. 
We downsample each original video at 24 frames per-second (FPS) to obtain a video at 12 FPS, resulting in around 115 frames in total for each video input.
Since a static background image is not available for each video captured in this dataset, we simply mask out the background to be white using the provided foreground mask.
\textit{(B) AIST}:
we clip out several video clips from the original videos, each lasts around 6-10 seconds. 
We then downsample each original clip at 60 FPS to obtain a video at 12 FPS, resulting in around 80-120 frames in total for each video input.
To obtain the static background image for each input video, we set the color at each location of the background image to be the median value of this location across the whole video clip.
To remove the considerable amount of shadows contained in this dataset, we run Mask R-CNN detection to obtain the human segmentation of each frame, which is then composited with the static background image to obtain the final image.
\textit{(C) ZJU-MoCap}:
each video in this dataset lasts around 10 seconds. 
We downsample each original video at 24 frames per-second (FPS) to obtain a video at 12 FPS, resulting in around 150 frames in total for each video input.
The static background image is also obtained via the background image extraction as is done in AIST.

\end{document}